# CANCER PROGNOSIS PREDICTION USING BALANCED STRATIFIED SAMPLING

J.S.Saleema [1], N.Bhagawathi[2], S.Monica[2], P.Deepa Shenoy[2*], K.R.Venugopal[2] and L.M.Patnaik[3]

[1] Department of Computer Science, Christ University, Bangalore, India
[2] Dept. of CSE, University Visvesvaraya College of Engineering, Bangalore, India
[3] Indian Institute of Science, Bangalore, India

## ABSTRACT

*High accuracy in cancer prediction is important to improve the quality of the treatment and to improve the rate of survivability of patients. As the data volume is increasing rapidly in the healthcare research, the analytical challenge exists in double. The use of effective sampling technique in classification algorithms always yields good prediction accuracy. The SEER public use cancer database provides various prominent class labels for prognosis prediction. The main objective of this paper is to find the effect of sampling techniques in classifying the prognosis variable and propose an ideal sampling method based on the outcome of the experimentation. In the first phase of this work the traditional random sampling and stratified sampling techniques have been used. At the next level the balanced stratified sampling with variations as per the choice of the prognosis class labels have been tested. Much of the initial time has been focused on performing the pre-processing of the SEER data set. The classification model for experimentation has been built using the breast cancer, respiratory cancer and mixed cancer data sets with three traditional classifiers namely Decision Tree, Naïve Bayes and K-Nearest Neighbour. The three prognosis factors survival, stage and metastasis have been used as class labels for experimental comparisons. The results shows a steady increase in the prediction accuracy of balanced stratified model as the sample size increases, but the traditional approach fluctuates before the optimum results.*

## 1. INTRODUCTION

Data analytics and big data are the market trend today in academia, research institutes and industries. The solutions for all analytical problems are obtained majorly through machine learning and data mining techniques. The major challenge of data handling exists in healthcare, as the rate of increase in patience is proportional to the rate of population growth and life style changes.

Cancer is the second most common cause of death, exceeded only by heart disease. The mortality rates are above 50% of the incidence in African and Asian countries, but the incidence are more in western countries though the mortality is less. The Surveillance, Epidemiology, and End Results (SEER) Program of the National Cancer Institute in United States is a well known repository of cancer database that is being updated every year [1]. The repository contains data from 1973 – 2010. Many other countries have also come up with the maintenance of the repository through various registries, as the overall number of cancer cases rapidly increasing worldwide.

Classification based decision systems are popular in cancer diagnostics since last decade. The hidden knowledge extracted will help the physicians to treat the patients effectively and cure them at the earliest. Product based cancer prediction calculators are available in market and literature. Ankit Agrawal in [2] has proposed a lung cancer outcome calculator using data mining technique for SEER data set and provides assistance to the doctors with a tool for survival prediction. Though efficient pre-processing improves the prediction accuracy, it is widely seen in [3-8] that the choice of sampling techniques highly affects the results of classifier irrespective of its power.

Random sampling and stratified sampling are the two traditional techniques used for selecting the data for classification. Though the stratified sampling is balanced at the strata level, the same may not hold good for the overall samples. This paper follows a methodology that selects a balanced sample based on the class labels used for experiments. If the class label contains a minority class particularly for multivariate classes, then a ratio is fixed while selecting the samples. If the class label contains a very minimal value below a fixed threshold, then the same would be considered as missing and hence will be removed during pre-processing. The pre-processing includes null removal, correlation and information gain ratio based removal of attributes. Few attributes are transformed and modified as required for best results. Some of the attributes have been removed in consultation with the domain experts and references [3].

The organization of the paper is as follows: section 2 addresses the background work related to the issues in sampling and other pre-processing challenges for improving the classifiers. Section 3 provides the data sets used for the experiments and the detailed methodology for building the proposed sampling model. Section 4 presents the empirical results analysis. Concluding remarks are given in section 5 along with a plan for the future work.

## 2. LITERATURE REVIEW

A mature discussion and proof on SEER data has been found in the literature on SEER statistics and in the public domain [1-3]. Few of the researchers are using the SEER*Stat software provided by SEER through NCI for cancer updates. The response variables for cancer prediction in this paper have been detected using the prominent variable identification study by Ankit Agrawal in [2, 3]. Various types of classification methods with different kinds of pre-processing techniques have been used for the individual cancer types like lung, colorectal and breast. Principle steps of SEER data pre-process contain the data conversion, merging or splitting the data based on the data types or the properties of the multivariate feature and construction of new feature [3]. Most of the researches considered survival prediction as a major problem. Lung cancer, breast cancer, colon cancer and colorectal cancer survival prediction using the base classifiers have been found in the literature.

Survival prediction of cancer patients are of two types. First one is based on the severity of diseases from the date of diagnosis; a patient's survival period is defined. The second one predicts using the follow-up features after the diagnosis. Less work is found in the survival of combined data where as there would be similarities in age, morphological data, stage and extent of disease that lead as a motivation for this paper. The SEER data set has been generated from different sources and the data the recoding strategy based on the medical dictionary among the SEER community leaves data sparse and skewed. The data is skewed mainly because of the imbalanced class labels that are used for prediction analysis as discussed in [4].

Class imbalance is treatable, but the way it has to be treated depends on the nature of the data set used and the choice of balancing technique. In [5] author tries to balance the samples with over sampling with replacement approach for balancing the minority class. Lemmens in [6] uses a weighted ratio balancing approach to predict churn and Liu explores the under sampling using the ensemble sampling approach as a solution for under sampling [7]. In this paper a simple balanced stratified sampling with ratio of samples based on the prediction class labels has been proposed. Importance of balanced sampling for improving the classification accuracy is found in Kahlilias work in [8].

The SEER data pre-processing and selection of prominent labels has been applied similar to the earlier work of this team as provided in [9]. The missing handling was complex during the initial stage of the pre-processing and the random imputation for varied attributes of [10] lent a supporting hand in the process. The varied survivability study as in [11-13] for the breast, colorectal and colon cancer motivated us for the selection of two existing cancer data sets and one new generation of mixed type cancer by selecting samples from breast and respiratory cancers for this study.

Quinlan in [15] proposed a new c4.5 decision tree algorithm that overcomes the disadvantages of basic ID3. The authors in [16] have explained the three algorithms decision tree, Bayes and k-NN in a detailed manner. As the SEER data set is a combination of nominal and numerical data these three algorithms would be the right choice for building a classifier. Other well known algorithms ANN and SVM from [17] have not been selected because of its time consuming pre-processing of data using rapidminer.

## 3. METHODOLOGY

The first part of this section briefs the data pre-processing steps, followed by the sampling techniques and finally the classifiers with prediction and evaluation. Figure 1 presents the proposed framework with all the major phases as a separate block.

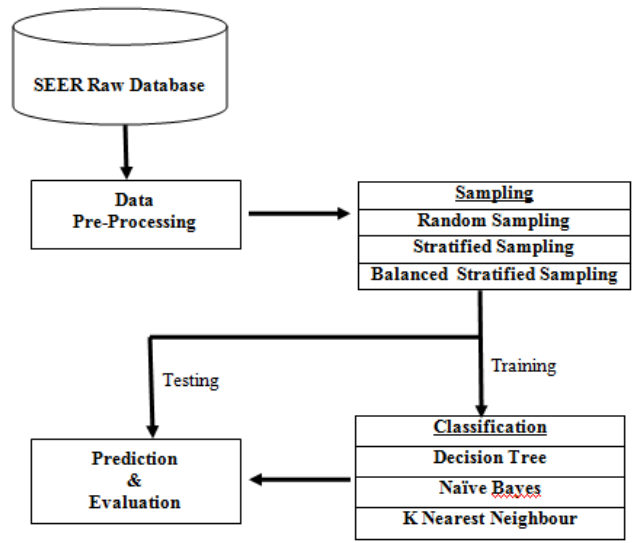

Figure 1. Framework of the proposed model

### 3.1. Data Set Description

SEER public use database from the program of National Cancer Institute (NCI) has been used for the empirical study of this paper [1]. SEER program currently collects data from approximately 26% of the USA population. Periodical incidences have been recorded as separate files and the incidence files of all eight cancer categories during the year 2000-2007 have been limited in this paper.

Each incidence text file contains different types of cancer data sets. More than two hundred thousands of records are available in a single file. A patient profile is a single record with 254 characters representing a total of 118 attributes. The attributes are both nominal and numerical in type [2]. Only the breast and respiratory raw files have been used in the first phase of pre-processing. A mixed file has been generated using the two types as a third file in the experiment. Figure 2 is a sample MATLAB GUI developed for generating a mixed data set from breast and respiratory cancer files. The repeated generation of samples from the source files have been merged in the excel repositories.

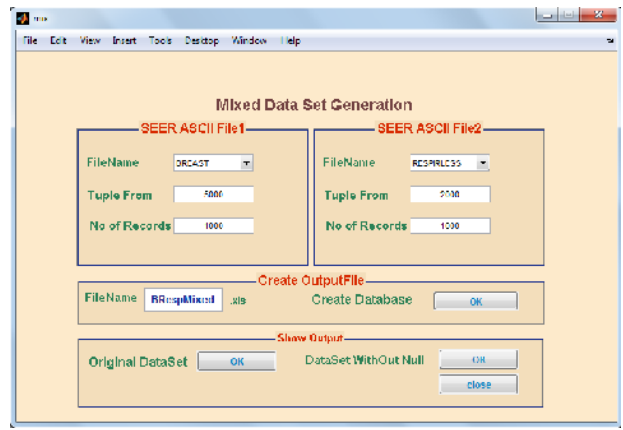

Figure 2. Sample frame of mixed data set generation

There exist 19 categories of attributes in the SEER database. From the 50000 processed samples, around 20000 sub samples have been used from breast and respiratory data sets for implementation. The major categories are demographic details, site specific details, disease extension, multiple primaries and cause of death. The categories are of type nominal and continuous in nature.

### 3.2. Data Pre-processing

The pre-processing has been performed in two stages. In the first stage, the raw data from the SEER text file has been parsed into tokens as per the available data dictionary. The pre-processing has been coded using the MATLAB. The data in the form of matrix has been preserved in the excel file for further processes.

In the second phase the features and the samples from the incidence year 2000-2007 have been analyzed for missing variables and records with null values. Based on the importance of data the missing features or the samples have been removed. Renaming and transformation has been performed for few attributes that were updated by the SEER registry. We reduced few attributes manually by studying the actual parameters and the class label. Some attributes have been re-coded based on the SEER dictionary changes with respect to years. The attributes size has been reduced to 64 from 118 in this phase.

SEER provides the numeric variable and its relevant nominal recoded variables like age-recode, Histology recode, etc. We have used the re-coded nominal variables for the experiments. Survival variable has been re-coded using Survival Time Recode (STR) in terms of month, as the data is a combination of year and month. For example, if the input is 0211, the same has been converted to 33 months (02X12+11). As per the literature in [2] using Vital Status Recode (VSR) and Cause of Death (COD), the survival data has been converted into a binary class label as survived and not survived for further study in sampling.

As the target group of our study is between 2000 and 2007, based on ICD-0-2 or ICD-0-3 medical standards, few attributes have been changed from 2004 onwards. Such attributes have been identified and merged using the common definitions available from SEER dictionary. For generating the metastasis variable the Extent of Disease (EOD) has been merged with Collaborative Stage (CS).

In the third phase all the remaining attributes have been processed for final feature extraction. The correlated attributes from the same category and attributes with low information gain have been removed as discussed in [9] and reduced to 37 attributes. The operators' correlation and filtration from rapidminer have been used for this feature extraction with its default parameters. All filtered data sets have been stored in a separate excel file.

### 3.3. Sampling

The two traditional sampling techniques and the balanced stratified sampling have been used in this phase. The number of classes for the class labels metastasis, stage and survival are 10, 4 and 2 respectively. The implementation of sampling selection and classification has been performed in rapidminer tool [14].

#### 3.3.1. Random Sampling

A simple random sampling is the probability of getting equal chance of being selected for each and every tuple of a data set $D$. The choice of sub-sample $n$ from the selected data set $N\epsilon D$ ranging from 500 to 30000 have been fixed for the remaining sampling techniques. In this approach the $n$ is random. Direct random sample operator has been used from rapidminer for this analysis.

#### 3.3.2. Stratified Sampling

A stratified sampling is the formation of distinct strata of $m$ homogenous classes and the process of selecting the samples from each strata based on the required subsample $n$ from the $N$ total sample and $m$ strata. Random choice of numbers from each strata increase the occurrence of all class representatives but can also result in over or under-sampling based on the availability of tuples in each strata. The stratified sampling of rapid miner uses the maximum possible size of data from each stratum. There is no guarantee that each class owns a representative in this. It depends on the overall population of 50000 that has been extracted during the pre-processing stage.

### 3.3.3. Balanced Stratified Sampling

The over sampling, under sampling and imbalanced sampling have been ignored. Balanced stratified sampling is selecting the samples from *m* strata, but of equal size. If there are minority or majority classes then over-sampling or under-sampling respectively need to be performed as required based on the distribution of the classes. In this paper the first option of balancing is through ignoring the minority classes for the ratios above 1:100. As the response variable is of multivariate in nature, ignoring the minority class leads to minimal error. The second method adopted in this paper is to take equal size of samples from each strata by reducing the stratum to *p<m*, such that *p* changing as the size of the sub-sample changes gradually from 500 to 30000.

The balancing criteria has been maintained in an excel file with the required size (i.e from 500–30000) and that file has been given as input for the already built-in stratified model. As the survival attribute was in binary form, it was easy to pick the sample representatives. The second option of equiv-width has been used. For minority multivariate classes like stage and metastasis the ratio of 1:100 has been verified using excel filter and this was given as an input to the rapidminer for classification. This method is a fixed allocation of balanced classes in the literature of probability and statistics.

## 3.4. Classifiers

There are many techniques available in literature to predict and classify cancer patterns. Machine learning and ANN systems are widely used in biomedical fields for various applications. Review of literature shows that applications of these learning approaches have improved the accuracy of cancer classification and survival prediction when compared to the other statistical methods. But the system must be exercised to enhance confidence, quality and reliability of reported data [12]. To test the variations of sampling techniques three basic classifiers have been chosen for the implementation of this paper. The algorithms are famous for nominal class labels [15-17].

### 3.4.1. Decision Tree

In decision tree classification the construction of tree starts by identifying an attribute to create a root node based on the information theory concept of entropy. Each such node splits the tree until a termination condition is met. The tree thus generated using the 36 attributes and one class labels will help the model to predict and evaluate the prediction accuracy for further testing. As the input attributes are of nominal and numeric type, the decision tree that works similar to Quinlan's c4.5 [15] or CART has been selected among the various algorithms available in rapidminer. The default information gain settings have been used for training the decision tree. Initial trial was performed using the ID2 simple decision tree in rapidminer. The data conversion process for the numerical variables was time consuming and hence the c4.5 has been decided for our complete analysis.

### 3.4.2. Naïve Bayes

Naïve Bayes works using the prior knowledge from the tuple that helps in predicting the future with likelihood estimation: A Naive Bayes classifier is a simple and low cost probabilistic classifier works on the Bayes theorem. A Bayesian approach divides the posterior probability into a prior distribution and a likelihood function. In "(1)" Probability of *X* given *C* is calculated using the probability of *C* given *X* and the prior probability of *X* and *C*. The outcome of (1) will be used to predict the unknown class *C* for any given *X* using *P(C/X)*.

$$P(X/C) = \frac{P(C/X)P(X)}{P(C)} \qquad (1)$$

The X vectors with 36 features and one class label C from survival, stage and metastasis have been used for training the algorithm. The 60:40 ratios have been picked using the rapidminer split operator for random and stratified sampling. Manual filtering form excel file has been used for balanced sampling. The procedure has been iteratively executed and the experiment results have been stored in a separate file for comparisons.

### 3.4.3. K-Nearest Neighbours

K Nearest Neighbours is a method that finds its closest neighbour from the training sample space that is having high similarity with a member in testing sample. Distance measures are primarily used to find the k-nearest neighbour (k-NN). The k-NN is a type of instance-based learning, or lazy learning, where the function is only approximated locally and all computation is deferred until classification. The default choice of *k=1* was initially used for testing the working process of the model. Later *k=5* and *k=10* have been used for variation analysis. It is found that the accuracy was less deviated during this test and thus kept the *k* value as *10* for further analysis. The rapidminer k-NN operator with the default mixed distance measures have been used, as the input variables are of numeric and nominal in nature.

## 4. EXPERIMENTAL RESULTS

The SEER incidence raw data files of the year 2000-2007 have been used for the implementation. MATLAB code has been developed for the pre-processing phase to parse, transform and reduce the attributes. Based on [4], few of the important features like stage, grade, marital status, etc have been converted from numerical to nominal. Few like survival time recode and sequence number have been split and reconstructed. As the source file contains above 200000 records, parsing of records has been performed in batches of 50000 each. The samples are drawn from different batches and collated into one for further processing. Table 1 presents the combination of size of samples, classifier algorithms and the caner types used for the comparison.

Table 1. Implementation features

| Sample size | Classifiers | Data Set | Training: Testing |
|---|---|---|---|
| 500 | Naïve Bayes | Breast Cancer | 60:40 |
| 1000 | | | |
| 2000 | | | |
| 5000 | Decision Tree | Respiratory Cancer | |
| 10000 | | | |
| 15000 | | | |
| 20000 | KNN | Mixed Type | |
| 25000 | | | |
| 30000 | | | |

Training and testing data split is performed with 60:40 ratios as this is the default best split. Using the split operator of rapidminer this has been experimented for random and stratified sampling. Minimum of 10 iterations have been performed for each sample size and the best of the ten results

15

were recorded here as output. The 10-fold cross validation was not performed because of the difficulty in selecting the balanced stratified samples as per our proposed definition. The time factor is a limitation of our work with could be resolved in further studies in the offing.

The processed excel file has given as an input to the rapidminer tool for classification. The result of all sampling techniques used for different class labels with various classifiers are discussed below in detail. Accuracy of the classifier has been kept as the evaluation measure for this paper. Table 2 shows the simple random and stratified sample output of the breast cancer data set for all three classifiers with stage as class label.

The result of the balanced sample for the same data set as that of table 2 is shown in table 3. It has been observed the balanced sampling technique gradually improves the accuracy and also show a good difference in performance. Similar experiment has been conducted for all other class labels for the scaling size of samples. The metastasis is a generated class from the severity of the disease using the increased primary node and multiple location of disease extent. It is believed that the balanced stage scores maximum accuracy out of all other members. The result is consistent with the other two classifiers too.

Table 2. Breast cancer performance for class stage using stratified and random sampling

| Sample Size | Stratified | | | Random | | |
|---|---|---|---|---|---|---|
| | DT | NB | KNN | DT | NB | KNN |
| 500 | 80.90% | 84.92% | 51.76% | 85.00% | 85.50% | 43.50% |
| 1000 | 66.83% | 90.02% | 49.88% | 69.58% | 89.03% | 50.37% |
| 2000 | 84.61% | 93.12% | 52.57% | 84.25% | 91.62% | 51.38% |
| 5000 | 66.87% | 94.85% | 55.62% | 85.29% | 93.25% | 57.78% |
| 10000 | **90.75%** | **95.52%** | 58.32% | 87.55% | 95.30% | 59.65% |
| 15000 | 66.90% | 94.70% | 61.33% | 67.28% | 95.03% | 61.05% |
| 20000 | 84.20% | 93.95% | 61.42% | **85.38%** | 94.94% | **63.41%** |
| 25000 | 84.49% | 94.63% | 63.00% | 84.83% | 94.17% | 63.18% |
| 30000 | 84.72% | 94.73% | **63.73%** | 84.37% | **95.84%** | 64.14% |

Table 3. Breast cancer performance for class stage using balanced stratified sampling

| Sample Size | Balanced | | |
|---|---|---|---|
| | DT | NB | KNN |
| 500 | 88.50% | 67.60% | 42.00% |
| 1000 | 91.50% | 69.00% | 47.25% |
| 2000 | 92.50% | 71.33% | 47.50% |
| 5000 | 93.50% | 72.10% | 54.75% |
| 10000 | 96.65% | 73.62% | 62.38% |
| 15000 | 96.82% | 76.15% | 64.88% |
| 20000 | 97.33% | 78.00% | 66.91% |
| 25000 | 98.20% | 78.25% | 69.00% |
| 30000 | **98.40%** | **79.00%** | **69.49%** |

As the stratified samples are more or less closer to the result, only the comparison of random and balanced sampling graph for all the three class labels for the respiratory data set has been presented in figure 3. The result presented in this is specific to KNN, as this was the least performer in breast cancer.

The result of the mixed cancer data set for the stage and metastasis is shown in figure 4 for further discussion ahead. As the survival has been studied in detail by many authors in the literature, only the stage and metastasis is shown in this figure. It is observe that the accuracy of the balanced stage and metastasis are almost equal throughout in the mixed type. It is also observed that the decision tree performed well in all cases of our experiment.

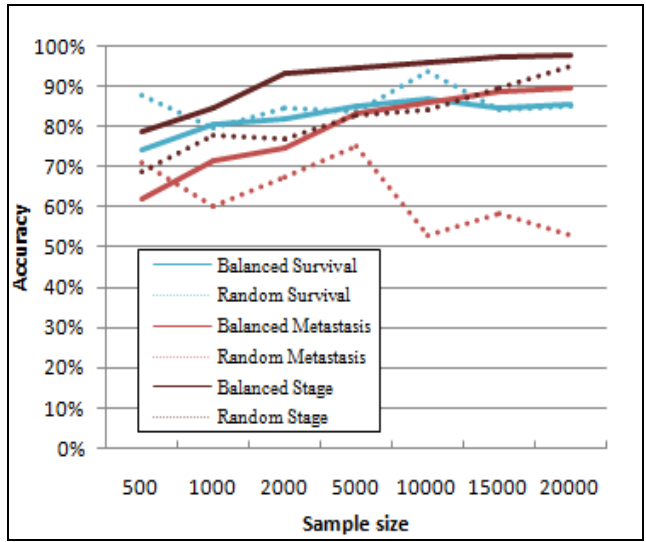

Figure 3. Performance of Respiratory Cancer for KNN Classifier

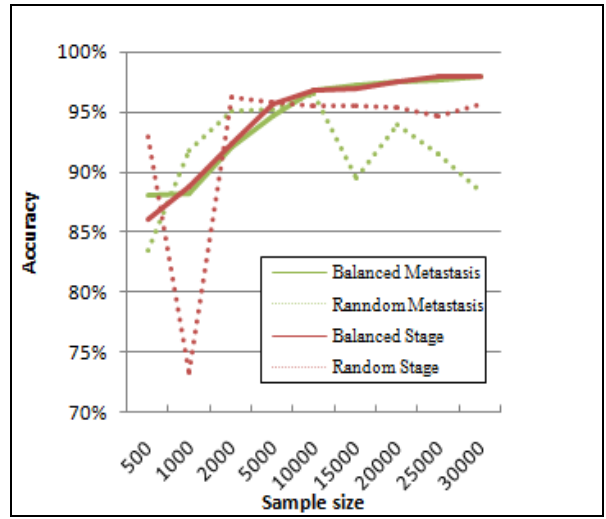

Figure 4. Performance of Mixed Cancer for Decision Tree Classifier

## 5. CONCLUSION

This paper is about predicting survival, metastasis, or stage using various techniques such as Naive Bayesian, decision tree and k-NN algorithm. As the SEER data size is very large, ample pre-processing and proper sampling need to be used while modelling the classifier. The objective of this paper is to compare three sampling techniques: random, stratified, and balanced stratified. The model has been tested with the SEER data sets and it shows a commendable variance in the performance. It has been observed that the balanced sampling always shows incremental performance proportional to the size of the samples. The experiment results shows a steady increase in the prediction accuracy of balanced stratified model as the sample size increases, but the traditional approach fluctuates before the optimum results. The work will be continued in future to study the performance of the modified balanced approach for an ensemble classifiers and also to automate the selection of balanced samples.

## REFERENCES

[1] SEER Publication, Cancer Facts, Surveillance Research Program, Cancer Statistics Branch, limited-use data (1973-2007). Available at: http:// seer.cancer.gov/data/.

[2] A. Agrawal, S. Misra, R. Narayanan, L. Polepeddi, and Alok Choudhary, "A lung cancer outcome calculator using ensemble data mining on SEER data, " Proceedings of the ACM SIGKDD International Conference on Knowledge Discovery and Data Mining. August 2011

[3] R. Al-Bahrani, A. Agrawal, and A. Choudhary, "Colon cancer survival prediction using ensemble data mining on SEER data," in Proceedings of the IEEE Big Data Workshop on Bioinformatics and Health Informatics (BHI), 2013.

[4] S. Li, Z. Wang, G. Zhou, and S. Y. Mei Lee, "Semi-supervised learning for imbalanced sentiment classification," in Proceedings of the Twenty-Second international joint conference on Artificial Intelligence. AAAI Press. vol. 3, pp. 1826-1831, 2011.

[5] J. Thongkam, G. Xu, Y. Zhang, and F. Huang, "Toward breast cancer survivability prediction models through improving training space," Expert Systems with Applications.vol 36(10), pp. 12200-12209, 2009.

[6] Lemmens, Aurélie and C. Croux, "Bagging and boosting classification trees to predict churn," Journal of Marketing Research. pp. 276-286, 2006.

[7] X. Liu, J. Wu, and Z. Zhou, "Exploratory under-sampling for class-imbalance learning," IEEE Transactions on Systems, Man, and Cybernetics. vol. 39(2), pp.539-550, 2009.

[8] M. Khalilia, S. Chakraborty, and M. Popescu, "Predicting disease risks from highly imbalanced data using random forest," BMC Medical Informatics and Decision Making. 2011.

[9] J. S. Saleema, B. Sairam, S. D. Naveen, K. Yuvaraj and P Deepa Shenoy, "Prominent label identification and multi-label classification for cancer prognosis prediction," TENCON 2012 - 2012 IEEE Region 10 Conference. Cebu. November 2012.

[10] J. Chen, J. N. K. Rao And R. R. Sitter, "Efficient random imputation for missing data in complex surveys," Statistica Sinica, vol. 10, pp 1153-1169, 2000.

[11] A. Bellaachia, E. Guven, "Predicting breast cancer survivability using data mining techniques," Age: Omaha, vol. 58, pp. 110-113, 2000.

[12] S. Kassem Fathy, "A prediction survival model for colorectal cancer," Proceedings of the 2011 American conference on applied mathematics and the 5th WSEAS international conference on Computer engineering and applications, pp 36-42, 2011.

[13] F.E Ahmed, "Artificial neural network for diagnosis and survival prediction in colon cancer," Molecular Cancer, vol. 4, no.29, 2005.

[14] Rapid Miner: Community Edition, Data Mining Software, Available at: http://rapidminer.com/ products/rapidminer-studio/.

[15] J. R. Quinlan, "Induction of decision trees," Machine Learning, vol. 1, pp. 81-106, 1986.

[16] J. Han and M. Kamber, Data Mining Concepts and Techniques. San Francisco, CA: Morgan Kaufmann, 2000.

[17] E. Alpaydin, Introduction to machine learning, 2nd ed. New Delhi: Prentice-Hall, 2010.